# NCBO Ontology Recommender 2.0: An Enhanced Approach for Biomedical Ontology Recommendation

Marcos Martínez-Romero[1,*] <marcosmr@stanford.edu>, Clement Jonquet[1,3] <jonquet@lirmm.fr>, Martin J. O'Connor[1] <martin.oconnor@stanford.edu>, John Graybeal[1] <jgraybeal@stanford.edu>
Alejandro Pazos[2] <apazos@udc.es>, Mark A. Musen[1] <musen@stanford.edu>

[1] Stanford Center for Biomedical Informatics Research, 1265 Welch Road, Stanford University School of Medicine, Stanford, CA 94305-5479, USA.

[2] Department of Information and Communication Technologies, Computer Science Building, Elviña Campus, University of A Coruña, 15071 A Coruña, Spain.

[3] Laboratory of Informatics, Robotics and Microelectronics of Montpellier (LIRMM), University of Montpellier, 161 rue Ada, 34095 Montpellier Cdx 5, France.

* Corresponding author

# Abstract

**Background.** Ontologies and controlled terminologies have become increasingly important in biomedical research. Researchers use ontologies to annotate their data with ontology terms, enabling better data integration and interoperability across disparate datasets. However, the number, variety and complexity of current biomedical ontologies make it cumbersome for researchers to determine which ones to reuse for their specific needs. To overcome this problem, in 2010 the National Center for Biomedical Ontology (NCBO) released the Ontology Recommender, which is a service that receives a biomedical text corpus or a list of keywords and suggests ontologies appropriate for referencing the indicated terms.

**Methods.** We developed a new version of the NCBO Ontology Recommender. Called Ontology Recommender 2.0, it uses a novel recommendation approach that evaluates the relevance of an ontology to biomedical text data according to four different criteria: (1) the extent to which the ontology covers the input data; (2) the acceptance of the ontology in the biomedical community; (3) the level of detail of the ontology classes that cover the input data; and (4) the specialization of the ontology to the domain of the input data.

**Results.** Our evaluation shows that the enhanced recommender provides higher quality suggestions than the original approach, providing better coverage of the input data, more detailed information about their concepts, increased specialization for the domain of the input data, and greater acceptance and use in the community. In addition, it provides users with more explanatory information, along with suggestions of not only individual ontologies but also groups of ontologies to use together. It also can be customized to fit the needs of different ontology recommendation scenarios.

**Conclusions.** Ontology Recommender 2.0 suggests relevant ontologies for annotating biomedical text data. It combines the strengths of its predecessor with a range of adjustments and new features that improve its reliability and usefulness. Ontology Recommender 2.0 recommends over 500 biomedical ontologies from the NCBO BioPortal platform, where it is openly available (both via the user interface at http://bioportal.bioontology.org/recommender, and via a Web service API).

# 1 Background

During the last two decades, the biomedical community has grown progressively more interested in ontologies. Ontologies provide the common terminology necessary for biomedical researchers to describe their datasets, enabling better data integration and interoperability, and therefore facilitating translational discoveries [1, 2].

BioPortal [3, 4], developed by the National Center for Biomedical Ontology (NCBO) [5], is a highly used platform[1] for hosting and sharing biomedical ontologies. BioPortal users can publish their ontologies as well as submit new versions. They can browse, search, review, and comment on ontologies, both interactively through a Web interface, and programmatically via Web services. In 2008, BioPortal[2] contained 72 ontologies and 300.000 ontology classes. As of 2017, the number of ontologies exceeds 500, with more than 7.8 million classes, making it one of the largest public repositories of biomedical ontologies.

The great number, complexity, and variety of ontologies in the biomedical field present a challenge for researchers: how to identify those ontologies that are most relevant for annotating, mining or indexing particular datasets. To address this problem, in 2010 the NCBO released the first version of its Ontology Recommender (henceforth 'Ontology Recommender 1.0' or 'original Ontology Recommender') [6], which informed the user of the most appropriate ontologies in BioPortal to annotate textual data. It was, to the best of our knowledge, the first biomedical ontology recommendation service, and it became widely known and used by the community[3]. However, the service has some limitations, and a significant amount of work has been done in the field of ontology recommendation since its release. This motivated us to analyze its weaknesses and to design a new recommendation approach.

---

[1] The BioPortal API received 18.8M calls/month on average in 2016. The BioPortal website received 306.9K pageviews/month on average in 2016 (see Additional file 1 for more detailed traffic data). The two main BioPortal papers [3, 4] accumulate 923 citations at the time of writing this paper, with 145 citations received in 2016.
[2] http://bioportal.bioontology.org/
[3] At the time of writing this paper, there are 63 citations to the NCBO Ontology Recommender 1.0 paper [6]. The Ontology Recommender 1.0 API received 7.1K calls/month on average in 2014. The Ontology Recommender webpage received 1.4K pageviews/month on average in 2014. Detailed traffic data is provided in Additional file 1.

The main contributions of this paper are the following:

1. A state-of-the-art approach for recommending biomedical ontologies. Our approach is based on evaluating the relevance of an ontology to biomedical text data according to four different criteria, namely: ontology coverage, ontology acceptance, ontology detail, and ontology specialization.

2. A new ontology recommendation system, the NCBO Ontology Recommender 2.0 (henceforth 'Ontology Recommender 2.0' or 'new Ontology Recommender'). This system has been implemented based on our approach, and it is openly available at BioPortal.

Our research is particularly relevant both for researchers and developers who need to identify the most appropriate ontologies for annotating textual data of biomedical nature (e.g., journal articles, clinical trial descriptions, metadata about microarray experiments, information on small molecules, electronic health records, etc.). Our ontology recommendation approach can be easily adapted to other domains, as it will be illustrated in the Discussion section. Overall, this work advances prior research in the fields of ontology evaluation and recommendation, and provides the community with a useful service which is, to the best of our knowledge, the only ontology recommendation system currently available to the public.

## 1.1 Related work

Much theoretical work has been done over the past two decades in the fields of ontology evaluation, selection, search, and recommendation. Ontology evaluation has been defined as the problem of assessing a given ontology from the point of view of a particular criterion, typically in order to determine which of several ontologies would best suit a particular purpose [7]. As a consequence, ontology recommendation is fundamentally an ontology evaluation task because it addresses the problem of evaluating and consequently selecting the most appropriate ontologies for a specific context or goal [8, 9].

Early contributions in the field of ontology evaluation date back to the early 1990s and were motivated by the necessity of having evaluation strategies to guide and improve the ontology engineering process [10–12]. Some years later, with the birth of the Semantic Web [13], the need for reusing ontologies across the Web motivated the development of the first ontology search engines [14–16], which made it possible to retrieve all ontologies satisfying some basic requirements. These engines usually returned only the ontologies that had the query term itself in their class or property names [17]. However, the process of recommending ontologies involves more than that. It is a complex process that comprises evaluating all candidate ontologies according to a variety of criteria, such as coverage, richness of the ontology structure [18–20], correctness, frequency of use [21], connectivity [18], formality, user ratings [22], and their suitability for the task at hand.

In biomedicine, the great number, size, and complexity of ontologies have motivated strategies to help researchers find the best ontologies to describe their datasets. Tan and Lambrix [23] proposed a theoretical framework for selecting the best ontology for a particular text-mining application and manually applied it to a gene-normalization task. Alani et al. [17] developed an ontology-search strategy that uses query-expansion techniques to find ontologies related to a particular domain (e.g., Anatomy). Maiga and Williams [24] conceived a semi-automatic tool that makes it possible to find the ontologies that best match a list of user-defined task requirements.

The most relevant alternative to the NCBO Ontology Recommender is BiOSS [21, 25], which was released in 2011 by some of the authors of this paper. BiOSS evaluates each candidate ontology according to three criteria: (1) the input coverage; (2) the semantic richness of the ontology for the input; and (3) the acceptance of the ontology. However, this system has some weaknesses that make it insufficient to satisfy many ontology reuse needs in biomedicine. BiOSS' ontology repository is not updated regularly, so it does not take into account the most recent revisions to biomedical ontologies. Also, BiOSS evaluates ontology acceptance by counting the number of mentions of the ontology name in Web 2.0 resources, such as Twitter and Wikipedia. However, this method is not always appropriate because a large number of mentions do not always correspond to a high level of acceptance by the community (e.g., an ontology may be "popular" on Twitter because of a high number of negative comments about it). Another drawback is that the input to BiOSS is limited to comma-delimited keywords; it is not possible to suggest ontologies to annotate raw text, which is a very common use case in biomedical informatics.

In this work, we have applied our previous experience in the development of the original Ontology Recommender and the BiOSS system to conceive a new approach for biomedical ontology recommendation. The new approach has been used to design and implement the Ontology Recommender 2.0. The new system combines the strengths of previous methods with a range of enhancements, including new recommendation strategies and the ability to handle new use cases. Because it is integrated within the NCBO BioPortal, this system works with a large corpus of current biomedical ontologies and can therefore be considered the most comprehensive biomedical ontology recommendation system developed to date.

Our recommendations for the choice of appropriate ontologies centers around the use of ontologies to perform annotation of textual data. We define annotation as a correspondence or relationship between a term and an ontology class that specifies the semantics of that term. For instance, an annotation might relate *leucocyte* in some text to a particular ontology class *leucocyte* in the Cell Ontology. The annotation process will also relate textual data such as *white blood cell* and *lymphocyte* to the class *leucocyte* in the Cell Ontology, via synonym and subsumption relationships, respectively.

## 1.2 Description of the original approach

The original NCBO Ontology Recommender supported two primary use cases: (1) corpus-based recommendation, and (2) keyword-based recommendation. In these scenarios, the system recommended appropriate ontologies from the BioPortal ontology repository to annotate a text corpus or a list of keywords, respectively.

The NCBO Ontology Recommender invoked the NCBO Annotator [26] to identify all annotations for the input data. The NCBO Annotator is a BioPortal service that annotates textual data with ontology classes. Then, the Ontology Recommender scored all BioPortal ontologies as a function of the number and relevance of the annotations found, and ranked the ontologies according to those scores. The first ontology in the ranking would be suggested as the most appropriate for the input data. The score for each ontology was calculated according to the following formula:[4]

$$score(o,t) = \frac{\sum(annotationScore(a) + 2 * hierarchyLevel(a))}{log10(|o|)} \quad \forall a \in annotations(o,t)$$

such that:

$$score(o,t) \in \mathbb{R} : score(o,t) \geq 0$$

$$annotationScore(a) = \begin{cases} 10 \text{ if } annotationType = PREF \\ 8 \text{ if } annotationType = SYN \end{cases}$$

$$hierarchyLevel(a) \in \mathbb{Z} : hierarchyLevel(a) \geq 0$$

Here $o$ is the ontology that is being evaluated; $t$ is the input text; $score(o, t)$ represents the relevance of the ontology $o$ for $t$; $annotationScore(a)$ is the score for the annotation $a$; $hierarchyLevel(a)$ is the position of the matched class in the ontology tree, such that 0 represents the root level; $|o|$ is the number of classes in $o$; and $annotations(o,t)$ is the list of annotations ($a$) performed with $o$ for $t$, returned by the NCBO Annotator.

The *annotationScore*(*a*) would depend on whether the annotation was achieved with a class 'preferred name' (*PREF*) or with a class synonym (*SYN*). A preferred name is the human readable label that the authors of the ontology suggested to be used when referring to the class (e.g., *vertebral column*), whereas synonyms are alternate names for the class (e.g., *spinal column*, *backbone*, *spine*). Each class in BioPortal has a single preferred name and it may have any number of synonyms. Because synonyms can be imprecise, this approach favored matches on preferred names.

---

[4] This formula is slightly different from the scoring method presented in the paper describing the original Ontology Recommender Web service [6]. It corresponds to an upgrade done in the recommendation algorithm in December 2011, when BioPortal 3.5 was released, for which description and methodology was never published. The normalization strategy was improved by applying a logarithmic transformation to the ontology size to avoid a negative effect on very large ontologies. Mappings between ontologies, used to favor reference ontologies, were discarded due to the small number of manually created and curated mappings that could be used for such a purpose. The hierarchy-based semantic expansion was replaced by the position of the matched class in the ontology hierarchy.

The normalization by ontology size was intended to discriminate between large ontologies that offer good coverage of the input data, and small ontologies with both correct coverage and better specialization for the input data's domain. The granularities of the matched classes (i.e., *hierarchyLevel*(*a*)) were also considered, so that annotations performed with granular classes (e.g., *epithelial cell proliferation*) would receive higher scores than those performed with more abstract classes (e.g., *biological process*).

For example, Table 1 shows the top five suggestions of the original Ontology Recommender for the text *Melanoma is a malignant tumor of melanocytes which are found predominantly in skin but also in the bowel and the eye.* In this example, the system considered that the best ontology for the input data is the National Cancer Institute Thesaurus (NCIT).

**Table 1. Ontologies suggested by the original Ontology Recommender for the sample input text *Melanoma is a malignant tumor of melanocytes which are found predominantly in skin but also in the bowel and the eye.*** For each ontology, the table shows its position in the ranking, the acronym of the ontology in BioPortal[5], the number of annotations returned by the NCBO Annotator for the sample input, the terms annotated (or 'covered') by those annotations and the ontology score.

| Rank | Ontology | No. annotations | Terms annotated | Score |
| --- | --- | --- | --- | --- |
| 1 | NCIT | 21 | *melanoma, malignant tumor, melanocytes, found, skin, bowel, eye* | 55.2 |
| 2 | EHDA | 15 | *skin, eye* | 38.3 |
| 3 | EFO | 10 | *melanoma, malignant tumor, skin, bowel, eye* | 35.9 |
| 4 | LOINC | 18 | *melanoma, malignant, tumor, skin, bowel, eye* | 35.9 |
| 5 | MP | 9 | *melanoma, skin, bowel, eye* | 34.8 |

In the following sections, we summarize the most relevant shortcomings of the original approach, addressing input coverage, coverage of multi-word terms, input types and output information.

### 1.2.1 Input coverage

Input coverage refers to the fraction of input data that is annotated with ontology classes. Given that the goal is to find the best ontologies to annotate the user's data, high input coverage is the main requirement for ontology-recommendation systems. One of the shortcomings of the original approach is that it did not ensure that ontologies that provide high input coverage were ranked higher than ontologies with lower coverage. The approach was strongly based on the total number of annotations returned by the NCBO Annotator. However, a large number of annotations does not always imply high coverage. Ontologies with low input coverage can contain a great many classes that match only a few input terms, or match many repeated terms in a large text corpus.

In the previous example (see Table 1), EHDA (Human Developmental Anatomy Ontology) was ranked at the second position. However, it covers only two input terms: *skin* and *eye*. Clearly, it is not an appropriate ontology to annotate the input when compared with LOINC or EFO, which have almost three times more terms covered. The reason that EHDA was assigned a high score is that it contains 11 different *eye* classes (e.g., EHDA:4732, EHDA:3808, EHDA:5701) and 4 different *skin* classes (e.g., EHDA:6531, EHDA:6530, EHDA:7501), which provide a total of 15 annotations. Since the recommendation score computed using the original approach is directly influenced by the number of annotations, EHDA obtains a high relevance score and thus the second position in the ranking. This issue was also identified by López-García *et al*. in their study of the efficiency of automatic summarization techniques [27]. These authors noticed that EHDA was the most recommended ontology for a broad range of topics that the ontology actually did not cover well.

---

[5] See "List of abbreviations".

### 1.2.2 Multi-word terms

Biomedical texts frequently contain terms composed of several words, such as *distinctive arrangement of microtubules*, or *dental disclosing preparation*. Annotating a multi-word phrase or multi-word keyword with an ontological class that completely represents its semantics is a much better choice than annotating each word separately. The original recommendation approach was not designed to select the longest matches and consequently the results were affected.

As an example, Table 2 shows the top 5 ontologies suggested by the original Ontology Recommender for the phrase *embryonic cardiac structure*. Ideally, the first ontology in the ranking (SWEET) would contain the class *embryonic cardiac structure*. However, the SWEET ontology covers only the term *structure*. This ontology was ranked at the first position because it contains 3 classes matching the term *structure* and also because it is a small ontology (4549 classes).

**Table 2. Top 5 ontologies suggested by Ontology Recommender 1.0 for the sample input text *embryonic cardiac structure*.** For each ontology, the table shows its position in the ranking, the acronym of the ontology in BioPortal, the number of annotations returned by the NCBO Annotator for the sample input, the terms annotated (or 'covered') by those annotations and the ontology score.

| Rank | Ontology | No. annotations | Score | Terms covered |
|------|----------|-----------------|-------|---------------|
| 1 | SWEET | 3 | 13.7 | *structure* |
| 2 | NCIT | 4 | 10.5 | *embryonic, cardiac, structure* |
| 3 | HUPSON | 2 | 10.1 | *cardiac, structure* |
| 4 | VSO | 1 | 9.8 | *structure* |
| 5 | SNOMEDCT | 2 | 8.9 | *embryonic cardiac structure* |

Furthermore, SNOMEDCT, which does contain a class that provides a precise representation of the input, was ranked in the 5th position. There are 3 other ontologies in BioPortal that contain the class *embryonic cardiac structure*: EP, BIOMODELS and FMA. However, they were ranked 8, 11 and 32, respectively. The recommendation algorithm should assign a higher score to an annotation that covers all words in a multi-word term than it does to different annotations that cover all words separately.

### 1.2.3 Input types

Related work in ontology recommendation highlights the importance of addressing two different input types: text corpora and lists of keywords [28]. The original Ontology Recommender, while offering users the possibility of selecting among these two recommendation scenarios, would treat the input data in the same manner. To satisfy users' expectations, the system should process these two input types differently, to better reflect the information coded in the input about multi-word boundaries.

#### 1.2.4 Output information

The output provided by the original Ontology Recommender consisted of a list of ontologies ranked by relevance score. For each ontology, the Web-based user interface displayed the number of classes matched and the size of each recommended ontology. In contrast, the Web service could additionally return the particular classes matched in each ontology. This information proved insufficient to assure users that a recommended ontology was appropriate and better than the alternatives. For example, it was not possible to know what specific input terms were covered by each class. The system should provide enough detail both to reassure users, and to give them information about alternative ontologies.

In this section we have described the fundamental limitations of the original Ontology Recommender and suggested methods to address them. The strategy for evaluating input coverage must be improved. Additionally, there is a diversity of other recently-proposed evaluation techniques [8, 19, 25] that could enhance the original approach. Particularly, there are two evaluation criteria that could substantially improve the output provided by the system: (1) ontology acceptance, which represents the degree of acceptance of the ontology by the community; and (2) ontology detail, which refers to the level of detail of the classes that cover the input data.

## 2 Description of the new approach

In this section, we present our new approach to biomedical ontology recommendation. First, we describe our ontology evaluation criteria and explain how the recommendation process works. We then provide some implementation details and discuss improvements to the user interface.

The execution starts from the input data and a set of configuration settings. The NCBO Annotator [26] is then used to obtain all annotations for the input using BioPortal ontologies. Those ontologies that do not provide annotations for the input data are considered irrelevant and are ignored in further processing. The ontologies that provide annotations are evaluated one by one according to four evaluation criteria that address the following questions:

1. **Coverage:** To what extent does the ontology represent the input data?
2. **Acceptance:** How well-known and trusted is the ontology by the biomedical community?
3. **Detail:** How rich is the ontology representation for the input data?
4. **Specialization:** How specialized is the ontology to the domain of the input data?

According to our analysis of related work, these are the most relevant criteria for ontology recommendation. Note that other authors have referred to the *coverage* criterion as *term matching* [6], *class match measure* [19] and *topic coverage* [28]. *Acceptance* is related to *popularity* [21, 25, 28], because it measures the level of support provided to the ontology by the people in the community. Other criteria to measure ontology acceptance are *connectivity* [6], and *connectedness* [18], which assess the relevance of an ontology based on the number and quality of connections to an ontology by other ontologies. *Detail* is similar to *structure measure* [6], *semantic richness* [21, 25], *structure* [18], and *granularity* [24].

For each of these evaluation criteria, a score in the interval [0,1] is typically obtained. Then, all the scores for a given ontology are aggregated into a composite *relevance score*, also in the interval [0,1]. This score represents the appropriateness of that ontology to describe the input data. The individual scores are combined in accordance with the following expression:

$$score(o,t) = w_c * coverage(o,t) + w_a * acceptance(o) + w_d * detail(o,t) + w_s * specialization(o,t)$$

where $o$ is the ontology that is being evaluated, $t$ represents the input data, and $\{w_c, w_a, w_d, w_s\}$ are a set of predefined weights that are used to give more or less importance to each evaluation criterion, such that $w_c + w_a + w_d + w_s = 1$. Note that *acceptance* is the only criterion independent from the input data. Ultimately, the system returns a list of ontologies ranked according to their relevance scores.

## 2.1 Ontology evaluation criteria

The relevance score of each candidate ontology is calculated based on coverage, acceptance, detail, and specialization. We now describe these criteria in more detail.

### 2.1.1 Ontology coverage

It is crucial that ontology recommendation systems suggest ontologies that provide high coverage of the input data. As with the original approach, the new recommendation process is driven by the annotations provided by the NCBO Annotator, but the method used to evaluate the candidate ontologies is different. In the new algorithm, each annotation is assigned a score computed in accordance with the following expression:[6]

$$annotationScore2(a) = (annotationTypeScore(a) + multiWordScore(a)) * annotatedWords(a)$$

with:

$$annotationTypeScore(a) = \begin{cases} 10 & if\ annotationType = PREF \\ 5 & if\ annotationType = SYN \end{cases}$$

$$multiWordScore(a) = \begin{cases} 3 & if\ annotatedWords(a) > 1 \\ 0 & otherwise \end{cases}$$

In this expression, *annotationTypeScore*(*a*) is a score based on the annotation type which, as with the original approach, can be either 'PREF', if the annotation has been performed with a class preferred name, or 'SYN', if it has been performed with a class synonym. Our method assigns higher relevance to scores done with class preferred names than to those made with class synonyms because we have seen that many BioPortal ontologies contain synonyms that are not reliable (e.g., *Other variants* as a synonym of *Other Variants of Basaloid Follicular Neoplasm of the Mouse Skin* in the NCI Thesaurus).

The *multiWordScore*(*a*) score rewards multi-word annotations. It gives more importance to classes that annotate multi-word terms than to classes that annotate individual words separately (e.g., *blood*

---

[6] The function is called *annotationScore2* to differentiate it from the original *annotationScore* function.

*cell* versus *blood* and *cell*). Such classes better reflect the input data than do classes that represent isolated words.

The *annotatedWords*(*a*) function represents the number of words matched by the annotation (e.g., 2 for the term *blood cell*).

Sometimes, an ontology provides overlapping annotations for the same input data. For instance, the text *white blood cell* may be covered by two different classes, *white blood cell* and *blood cell*. In the original approach, ontologies with low input coverage were sometimes ranked among the top positions because they had multiple classes matching a few input terms, and all those annotations contributed to the final score. Our new approach addresses this issue. If an ontology provides several annotations for the same text fragment, only the annotation with the highest score is selected to contribute to the coverage score.

The coverage score for each ontology is computed as the sum of all the annotation scores, as follows:

$$coverage(o,t) = norm\left(\sum annotationScore2(a)\right) \quad \forall a \in selectedAnnotations(A)$$

where *A* is the set of annotations performed with the ontology *o* for the input *t*, *selectedAnnotations*(*A*) is the set of annotations that are left after discarding overlapping annotations, and *norm* is a function that normalizes the coverage score to the interval [0,1].

As an example, Table 3 shows the annotations performed with SNOMEDCT for the input *A thrombocyte is a kind of blood cell.* This example shows how our approach prioritizes (i.e., assigns a higher score to) annotations performed with preferred names over synonyms (e.g., *cell* over *entire cell*), and annotations performed with multi-word terms over single-word terms (e.g., *blood cell* over *blood* plus *cell*). The coverage score for SNOMEDCT would be calculated as 5+26=31, which would be normalized to the interval [0,1] by dividing it by the maximum coverage score. The maximum coverage score is obtained by adding the scores of all the annotations performed with all BioPortal ontologies, after discarding overlapping annotations.

**Table 3. SNOMEDCT annotations for the input *A thrombocyte is a kind of blood cell*.** The table shows the text fragment covered by each annotation, the name and type of the matched class, the annotation score, and the annotations selected to compute the relevance score for SNOMEDCT.

| Text | Matched class (type) | Annotation score | Selected |
|---|---|---|---|
| *thrombocyte* | *platelet* (SYN) | 5 | Yes |
| *blood cell* | *blood cell* (PREF) | (10+3)*2=26 | Yes |
| *blood* | *blood* (PREF) | 10 | No |
| *cell* | *cell structure* (SYN) | 5 | No |
| *cell* | *cell* (PREF) | 10 | No |
| *cell* | *entire cell* (SYN) | 5 | No |

It is important to note that this evaluation of ontology coverage takes into account term frequency. That is, matched terms with several occurrences are considered more relevant to the input data than terms that occur less frequently. If an ontology covers a term that appears several times in the input, its corresponding annotation score will be counted each time and the coverage score for the ontology accordingly will be higher. In addition, because we select only the matches with the highest score, the frequencies are not distorted by terms embedded in one another (e.g., *white blood cell* and *blood cell*).

Our approach accepts two input types: free text and comma-delimited keywords. For the keyword input type, only those annotations that cover all the words in a multi-word term are considered. Partial annotations are immediately discarded.

### 2.1.2 Ontology acceptance

In biomedicine, some ontologies have been developed and maintained by widely known institutions or research projects. The content of these ontologies is periodically curated, extensively used, and accepted by the community. Examples of broadly accepted ontologies are SNOMEDCT [29] and Gene Ontology [30]. Some ontologies uploaded to BioPortal may be relatively less reliable, however. They may contain incorrect or poor quality content or simply be insufficiently up to date. It is important that an ontology recommender be able to distinguish between ontologies that are accepted as trustworthy and those that are less so.

Our approach proposes to estimate the degree of acceptance of each ontology based of information extracted from ontology repositories or terminology systems. Widely used examples of these systems in biomedicine include BioPortal, the Unified Medical Language System (UMLS) [31], the OBO Foundry [32], Ontobee [33], the Ontology Lookup Service (OLS) [34], and Aber-OWL [35]. The calculation of ontology acceptance is based on two factors: (1) The presence or absence of the ontology in ontology repositories; and (2) the number of visits (*pageviews*) to the ontology in ontology repositories in a recent period of time (e.g., the last 6 months). This method takes into account changes in ontology acceptance over time. The acceptance score for each ontology is calculated as follows:

$$acceptance(o) = w_{presence} * presenceScore(o) + w_{visits} * visitsScore(o)$$

where:

- *presenceScore*(*o*) is a value in the interval [0,1] that represents the presence of the ontology in a predefined list of ontology repositories. It is calculated as follows:

$$presenceScore(o) = \sum_{i=1}^{n} w_{p_i} * presence_i(o)$$

where $w_{p_i}$ represents the weight assigned to the presence of the ontology in the repository *i*, with $\sum_{i=1}^{n} w_{p_i} = 1$, and:

$$presence_i(o) = \begin{cases} 1 & if\ o\ is\ present\ in\ repository\ i \\ 0 & otherwise \end{cases}$$

- *visitsScore*(*o*) represents the number of visits to the ontology on a given list of ontology repositories in a recent period of time. Note that this score can typically be calculated only for those repositories that are available on the Web and that have an independent page for each provided ontology. This score is calculated as follows:

$$visitsScore(o) = \sum_{i=1}^{n} w_{v_i} * visits_i(o)$$

where $w_{v_i}$ is the weight assigned to the ontology visits on the repository $i$, with $\sum_{i=1}^{n} w_{v_i} = 1$; $visits_i(o)$ represents the number of visits to the ontology in the repository $i$, normalized to the interval [0,1].

- $w_{presence}$ and $w_{visits}$ are weights that are used to give more or less importance each factor, with $w_{presence} + w_{visits} = 1$.

**Figure 1. Top 20 BioPortal ontologies according to their acceptance scores.** The x-axis shows the acceptance score in the interval [0, 100]. The y-axis shows the ontology acronyms. These acceptance scores were obtained by using UMLS to calculate the *presenceScore(o)*, BioPortal to compute the *visitsScore(o)*, and assigning the same weight to *pageviewsScore(o)* and *reposScore(o)* ($w_{pv}$=0.5, $w_{repos}$=0.5).

Figure 1 shows the top 20 accepted BioPortal ontologies according to our approach at the time of writing this paper. Estimating the acceptance of an ontology by the community is inherently subjective, but the above ranking shows that our approach provides reasonable results. All ontologies in the ranking are widely known and accepted biomedical ontologies that are used in a variety of projects and applications.

### 2.1.3 Ontology detail

Ontologies containing a richer representation for a specific input are potentially more useful to describe the input than less detailed ontologies. As an example, the class *melanoma* in the Human Disease Ontology contains a definition, two synonyms, and twelve properties. However, the class *melanoma* from the GALEN ontology does not contain any definition, synonyms, or properties. If a user needs an ontology to represent that concept, the Human Disease Ontology would probably be more useful than the GALEN ontology because of this additional information. An ontology recommender should be able to analyze the level of detail of the classes that cover the input data and to give more or less weight to the ontology according to the degree to which its classes have been specified.

We evaluate the richness of the ontology representation for the input data based on a simplification of the "semantic richness" metric used by BiOSS [25]. For each annotation selected during the coverage evaluation step, we calculate the detail score as follows:

$$detailScore(a) = \frac{definitionScore(a) + synonymsScore(a) + propertiesScore(a)}{3}$$

where *detailScore*(a) is a value in the interval [0,1] that represents the level of detail provided by the annotation *a*. This score is based on three functions that evaluate the detail of the knowledge representation according to the number of definitions, synonyms, and other properties of the matched class:

$$definitionScore(a) = \begin{cases} 1 \text{ if } |D| \geq k_d \\ |D|/k_d \text{ otherwise} \end{cases}$$

$$synonymsScore(a) = \begin{cases} 1 \text{ if } |S| \geq k_s \\ |S|/k_s \text{ otherwise} \end{cases}$$

$$propertiesScore(a) = \begin{cases} 1 \text{ if } |P| \geq k_p \\ |P|/k_p \text{ otherwise} \end{cases}$$

where |D|, |S| and |P| are the number of definitions, synonyms, and other properties of the matched class, and $k_d$, $k_s$ and $k_p$ are predefined constants that represent the number of definitions, synonyms, and other properties, respectively, necessary to get the maximum detail score. For example, using $k_s$=4 means that, if the class has 4 or more synonyms, then it will be assigned the maximum synonyms score, which would be 1. If it has fewer than 4 synonyms, for example 3, the synonyms score will be computed proportionally according to the expression above (i.e., 3/4). Finally, the detail for the ontology would be calculated as the sum of the detail scores of the annotations done with the ontology, normalized to [0,1]:

$$detail(o,t) = \frac{\sum detailScore(a)}{|A|} \quad \forall a \in selectedAnnotations(A)$$

Example: Suppose that, for the input *t = Penicillin is an antibiotic used to treat tonsillitis,* there are two ontologies O1 and O2 with the classes shown in Table 4.

**Table 4. Example of ontology classes for the input *Penicillin is an antibiotic used to treat tonsillitis*.** The table shows the ontology name, the class name, and the number of class definitions, synonyms and properties.

| Ontology | Class | No. definitions | No. synonyms | No. properties |
|---|---|---|---|---|
| O1 | *penicillin* | 1 | 2 | 7 |
|  | *antibiotic* | 1 | 7 | 16 |
| O2 | *penicillin* | 0 | 1 | 3 |
|  | *tonsillitis* | 0 | 0 | 2 |

Assuming that $k_d = 1$, $k_s = 4$ and $k_p = 10$, the detail score for O1 and O2 would be calculated as follows:

$$detail(O1, t) = \frac{\left(\frac{1 + 2/4 + 7/10}{3}\right) + \left(\frac{1+1+1}{3}\right)}{2} = 0.87$$

$$detail(O2, t) = \frac{\left(\frac{0 + 1/4 + 3/10}{3}\right) + \left(\frac{0+0+2/10}{3}\right)}{2} = 0.13$$

Given that O1 annotates the input with two classes that provide more detailed information than the classes from O2, the detail score for O1 is higher.

### 2.1.4 Ontology specialization

Some biomedical ontologies aim to represent detailed information about specific subdomains or particular tasks. Examples include the Ontology for Biomedical Investigations [36], the Human Disease Ontology [37] and the Biomedical Resource Ontology [38]. These ontologies are usually much smaller than more general ones, with only several hundred or a few thousand classes, but they provide comprehensive knowledge for their fields.

To evaluate ontology specialization, an ontology recommender needs to quantify the extent to which a candidate ontology fits the specialized nature of the input data. To do that, we reused the evaluation approach applied by the original Ontology Recommender, and adapted it to the new annotation scoring strategy. The specialization score for each candidate ontology is calculated according to the following expression:

$$specialization(o,t) = norm\left(\frac{\sum(annotationScore2(a) + 2 * hierarchyLevel(a))}{log_{10}(|o|)}\right) \forall a \in A$$

where *o* is the ontology being evaluated, *t* is the input text, *annotationScore2(a)* is the function that calculates the relevance score of an annotation (see Section 2.1.1), *hierarchyLevel(a)* returns the level of the matched class in the ontology hierarchy, and *A* is the set of all the annotations done with the ontology *o* for the input *t*. Unlike the coverage and detail criteria, which consider only *selectedAnnotations(A)*, the specialization criterion takes into account all the annotations returned by the Annotator (i.e., *A*). This is generally appropriate because an ontology that provides multiple annotations for a specific text fragment is likely to be more specialized for that text than an ontology that provides only one annotation for it. The normalization by ontology size aims to assign a higher score to smaller, more specialized ontologies. Applying a logarithmic function decreases the impact of ontologies with a very large size. Finally, the *norm* function normalizes the score to the interval [0,1].

**Table 5. Ontology size and annotation details for the ontologies in Table 4.** This table shows the number of classes (size) of each ontology, the class names, the annotation types, the annotation scores, and the level of each class in the ontology hierarchy, such that '1' corresponds to the root (or top) level, '2' correspond to the level below the root classes, '3' to the next level, and so on.

| Ontology | Size | Class | Annotation type | Annotation score | Hierarchy level |
|---|---|---|---|---|---|
| O1 | 120,000 | *penicillin* | PREF | 10 | 5 |
| | | *antibiotic* | SYN | 5 | 3 |
| O2 | 800 | *penicillin* | SYN | 5 | 6 |
| | | *tonsillitis* | PREF | 10 | 12 |

Using the same hypothetical ontologies, input, and annotations from the previous example, and taking into account the size and annotation details shown in Table 5, the specialization score for O1 and O2 would be calculated as follows:

$$specialization(O1,t) = norm\left(\frac{(10 + 2 * 5) + (5 + 2 * 3)}{log_{10}(120000)}\right) = norm\left(\frac{31}{5.08}\right) = norm(6.10)$$

$$specialization(O2, t) = norm\left(\frac{(5 + 2 * 6) + (10 + 2 * 12)}{log_{10}(800)}\right) = norm\left(\frac{51}{2.90}\right) = norm(17.59)$$

It is possible to see that the classes from O2 are located deeper in the hierarchy than are those from O1. Also, O2 is a much smaller ontology than O1. As a consequence, according to our ontology-specialization method, O2 would be considered more specialized for the input than O1, and would be assigned a higher specialization score.

## 2.2 Evaluation of ontology sets

When annotating a biomedical text corpus or a list of biomedical keywords, it is often difficult to identify a single ontology that covers all terms. In practice, it is more likely that several ontologies will jointly cover the input [8]. Suppose that a researcher needs to find the best ontologies for a list of biomedical terms. If there is not a single ontology that provides an acceptable coverage it should then evaluate different combinations of ontologies and return a ranked list of ontology sets that, together, provide higher coverage. For instance, in our previous example (*Penicillin is an antibiotic used to treat tonsillitis*), O1 covers the terms *penicillin* and *antibiotic* and O2 covers *penicillin* and *tonsillitis*. None of those ontologies provides full coverage of all the relevant input terms. However, by using O1 and O2 together, it is possible to cover *penicillin*, *antibiotic*, and *tonsillitis*.

Our method to evaluate ontology sets is based on the "ontology combinations" approach used by the BiOSS system [21]. The system generates all possible sets of 2 and 3 candidate ontologies (3 being the default maximum, though users may modify this limit according to their specific needs) and it evaluates them using the criteria presented previously. To improve performance, we use some heuristic optimizations to discard certain ontology sets without performing the full evaluation process for them. For example, a set containing two ontologies that cover exactly the same terms will be immediately discarded because that set's coverage will not be higher than that provided by each ontology individually.

The relevance score for each set of ontologies is calculated using the same approach as for single ontologies, in accordance with the following expression:

$$scoreSet(O,t) = w_c * coverageSet(O,t) + w_a * acceptanceSet(O) + w_d * detailSet(O,t) + w_s * specializationSet(O,t)$$

where $O = \{o \mid o$ is an ontology$\}$ and $|O| > 1$. The scores for the different evaluation criteria are calculated as follows:

- **coverageSet**: It is computed the same way as for a single ontology, but takes into account all the annotations performed with all the ontologies in the ontology set. The system selects the best annotations, and the set's input coverage is computed based on them.

- **acceptanceSet**, **detailSet**, and **specializationSet**: For each ontology, the system calculates its coverage contribution (as a percentage) to the set's coverage score. The recommender then uses this contribution to calculate all the other scores proportionally. By using this method, the impact (in terms of acceptance, detail and specialization) of a particular ontology on the set score will vary according to the coverage provided by such ontology.

## 2.3 Implementation details

Ontology Recommender 2.0 implements the ontology recommendation approach previously described in this paper. Figure 2 shows the architecture of Ontology Recommender 2.0. Like its predecessor, it has two interfaces: a Web service API[7], which makes it possible to invoke the recommender programmatically, and a Web-based user interface, which is included in the NCBO BioPortal[8].

---

[7] The API documentation is available at http://data.bioontology.org/documentation#nav_recommender
[8] The Web-based user interface is available at http://bioportal.bioontology.org/recommender

> **Figure 2. An overview of the architecture and workflow of Ontology Recommender 2.0.** (1) The input data and parameter settings are received through any of the system interfaces (i.e., Web service or Web UI), and are sent to the system's backend. (2) The evaluation process starts. The NCBO Annotator is invoked to retrieve all annotations for the input data. The system uses these annotations to evaluate BioPortal ontologies, one by one, according to four criteria: coverage, acceptance, detail and specialization. Because of the system's modular design, additional evaluation criteria can be easily added. The system uses BioPortal services to retrieve any additional information required by the evaluation process. For example, evaluation of ontology acceptance requires the number of visits to the ontology in BioPortal (*pageviews*), and checking whether the ontology is present in the Unified Medical Language System (UMLS) or not. Four independent evaluation scores are returned for each ontology (one per evaluation criterion). (3) The scores obtained are combined into a relevance score for the ontology. (4) The relevance scores are used to generate a ranked list of ontologies or ontology sets, which (5) is returned via the corresponding system's interface.

The Web-based user interface was developed using the Ruby-on-Rails Web framework and the Javascript language. Server side components were implemented using the Ruby language. These components interact with other BioPortal services to retrieve all the information needed to achieve the recommendation process.

The typical workflow is as follows. First, the Ontology Recommender calls the Annotator service to obtain all the annotations performed for the input data using all BioPortal ontologies. Second, for each ontology, it invokes other BioPortal services to obtain the number of classes in the ontology, the number of visits to each ontology in a recent period of time, and to check the presence of the ontology in UMLS. Third, for each annotation performed with the ontology, it makes several calls to retrieve the number of definitions, synonyms and properties of the ontology class involved in the annotation. The system has four independent evaluation modules that use all this information to assess each candidate ontology according to the four evaluation criteria proposed in our approach: coverage, acceptance, detail, and specialization. Because of the system's modular design, new ontology evaluation modules can be easily plugged in.

NCBO provides a Virtual Appliance for communities that want to use the Ontology Recommender locally. This appliance is a pre-installed copy of the NCBO software that users can run and maintain. More information about obtaining and installing the NCBO Virtual Appliance is available at the NCBO Wiki[9].

The system uses a set of predefined parameters to control how the different evaluation scores are calculated, weighted and aggregated. Given that high input coverage is the main requirement for ontology recommendation systems, the weight assigned by default to ontology coverage (0.55) is considerably higher than the weight assigned to ontology acceptance, detail and specialization (0.15). Our system uses the same coverage weight than the BiOSS system [21]. The default configuration provides appropriate results for general ontology recommendation scenarios. However, both the web interface and the REST service allow users to adapt the system to their specific needs by modifying the weights given to coverage, acceptance, knowledge detail, and specialization. The predefined values for all default parameters used by Ontology Recommender 2.0 are provided as an additional file [see Additional file 2].

Some Ontology Recommender users may need to obtain repeatable results over time. Currently, however, any changes in the BioPortal ontology repository, such as submitting a new ontology or removing an existing one, may change the suggestions returned by the Ontology Recommender for the same inputs. BioPortal services do not provide version-based ontology access, so services such as the Ontology Recommender and the Annotator always run against the latest versions of the ontologies. A possible way of dealing with this shortcoming would be to install the NCBO Virtual Appliance with a particular set of ontologies and keep them locally unaltered.

The Ontology Recommender 2.0 was released in August 2015, as part of BioPortal 4.20[10]. The traffic data for 2016 reflects the great interest of the community on the new system, with an average of 45.2K calls per month to the Ontology Recommender API, and 1.2K views per month on the Ontology Recommender webpage. These numbers represent an increase of more than 600% in the number of calls to the API over 2015, and more than 30% in the number of pageviews over 2015. Other widely used BioPortal services are Search, with an average of 873.9K calls per month to the API, and 72.9K pageviews per month in 2016; and the Annotator, with an average of 484.8K calls per month to the API, and 3K pageviews per month in 2016. Detailed traffic data for the Ontology Recommender and other top used BioPortal services for the period 2014-2016 is provided as an additional file [see Additional file 1]. The source code is available in GitHub[11] under a BSD License.

---

[9] https://www.bioontology.org/wiki/index.php/Category:NCBO_Virtual_Appliance
[10] BioPortal release notes: https://www.bioontology.org/wiki/index.php/BioPortal_Release_Notes

### 2.3.1 User Interface

Figure 3 shows the Ontology Recommender 2.0 user interface. The system supports two input types: plain text and comma-separated keywords. It also provides two kinds of output: ranked ontologies and ranked ontology sets. The advanced options section, which is initially hidden, allows the user to customize (1) the weights applied to the evaluation criteria, (2) the maximum number of ontologies in each set (when using the ontology sets output), and (3) the list of candidate ontologies to be evaluated.

**Figure 3. Ontology Recommender 2.0 user interface.** The user interface has buttons to select the input type (i.e., text or keywords) and output type (i.e., ontologies and ontology sets). A text area enables the user to enter the input data. The "Get Recommendations" button triggers the execution. The "advanced options" button shows additional settings to customize the recommendation process.

Figure 4 shows an example of the system's output when selecting "keywords" as input and "ontologies" as output. For each ontology in the output, the user interface shows its final score, the scores for the four evaluation criteria used, and the number of annotations performed with the ontology on the input. For instance, the most highly recommended ontology in Figure 4 is the Symptom Ontology (SYMP), which covers 17 of the 21 input keywords. By clicking on the different rows of the column "highlight annotations", the user can select any of the suggested ontologies and see which specific input terms are covered. Also, clicking on a particular term in the input reveals the details of the matched class in BioPortal. All scores are translated from the interval [0, 1] to [0, 100] for better readability. A score of '0' for a given ontology and evaluation criterion means that the ontology has obtained the lowest score compared to the rest of candidate ontologies. A score of '1' means that the ontology has obtained the highest score, in relation to all the other candidate ontologies.

---

[11] https://github.com/ncbo/ncbo_ontology_recommender

Figure 5 shows the "Ontology sets" output for the same keywords displayed in Figure 4. The output shows that using three ontologies (SYMP, SNOMEDCT and MEDDRA) it is possible to cover all the input keywords. Different colors for the input terms and for the recommended ontologies in Figure 5 distinguish the specific terms covered by each ontology in the selected set.

**Figure 4. Example of the "Ontologies" output.** The user interface shows the top recommended ontologies. For each ontology, it shows the position of the ontology in the ranking, the ontology acronym, the final recommendation score, the scores for each evaluation criteria (i.e., coverage, acceptance, detail, and specialization), and the number of annotations performed with the ontology. The "highlight annotations" button highlights the input terms covered by the ontology.

**Figure 5. Example of the "Ontology sets" output.** The user interface shows the top recommended ontology sets. For each set, it shows its position in the ranking, the acronyms of the ontologies that belong to it, the final recommendation score, the scores for each evaluation criteria (i.e., coverage, acceptance, detail, and specialization), and the number of annotations performed with all the ontologies in the ontology set. The "highlight annotations" button highlights the input terms covered by the ontology set.

### 2.3.2 Limitations

One of the shortcomings of the current implementation is that the acceptance score is calculated using data from only two platforms. BioPortal is used to calculate the visits score, and UMLS is used to calculate the presence score. There are other widely known ontology repositories that should be considered too. We believe that the reliability of the current implementation would be increased by taking into account *visits* and *presence* information from additional platforms, such as the OBO Foundry and the Ontology Lookup Service (OLS). Extending our implementation to make use of additional platforms would require us to have a consistent mechanism to check the presence of each candidate ontology into other platforms, as well as a way to access updated traffic data from them.

Another limitation is related to the ability to identify different variations of a particular term. The coverage evaluation metric is dependent on the annotations identified by the Annotator for the input data. The Annotator deals with synonyms and term inflections (e.g., *leukocyte, leukocytes, white blood cell*) by using the synonyms contained in the ontology for a particular term. For example, Medical Subject Headings (MeSH) provides 11 synonyms for the term *leukocytes*, including *leukocyte* and *white blood cells*. As a consequence, the Annotator would be able to perform an annotation between the input term *white blood cells* and the MESH term *leukocytes*. However, not all ontologies provide such level of detail for their classes, and therefore the Annotator may not be able to appropriately perform annotations with them. The NCBO, in collaboration with University of Montpellier, is currently investigating several NLP approaches to improve the Annotator service. Applying lemmatization to both the input terms and the dictionary used by the Annotator is one of the methods currently being tested. As soon as these new features will be made available in the Annotator, they will automatically be used by Ontology Recommender.

## 3   Evaluation

To evaluate our approach, we compared the performance of Ontology Recommender 2.0 to Ontology Recommender 1.0 using data from a variety of well-known public biomedical databases. Examples of these databases are PubMed, which contains bibliographic information for the fields of biomedicine and health; the Gene Expression Omnibus (GEO), which is a repository of gene expression data; and ClinicalTrials.gov, which is a registry of clinical trials. We used the API provided by the NCBO Resource Index[12] [39] to programmatically extract data from those databases.

## 3.1 Experiment 1: Input Coverage

We selected 12 widely known biomedical databases and extracted 600 biomedical texts from them, with 127 words on average, and 600 lists of biomedical keywords, with 17 keywords on average, producing a total of 1200 inputs (100 inputs per database). The databases used are listed in Table 6. Given the importance of input coverage, we first executed both systems for all inputs and compared the coverage provided by the top-ranked ontology. We focused on the top-ranked ontology because the majority of users always select the first result obtained [40]. The strategy we used to calculate the ontology coverage differed depending on the input type:

- For texts, the coverage was computed as the percentage of input words covered by the ontology with respect to the total number of words that could be covered using all BioPortal ontologies together.

- For keywords, the coverage was computed as the percentage of keywords covered by the ontology divided by the total number of keywords.

---

[12] The NCBO Resource Index is an ontology-based index that provides access to over 30 million biomedical records from 48 widely-known databases. It is available at: http://bioportal.bioontology.org/.

**Table 6. Databases used for experiment 1.** The table shows the database name, its acronym, the main topic of the database, the specific field from which the information was extracted, and the type of textual data extracted (i.e., text or keywords).

| Database name | Acronym | Topic | Source field | Type |
|---|---|---|---|---|
| ARRS GoldMiner | GM | Biomedical images | Image caption | Text |
| Autism Database (AutDB) | AUTDB | Autism spectrum disorders | Phenotype profile | Text |
| Gene Expression Omnibus | GEO | Gene expression | Summary | Text |
| Integrated Disease View | IDV | Disease and treatment | Description | Text |
| PubMed | PM | Biomedicine | Abstract | Text |
| PubMed Health Drugs | PMH | Drugs | Why is this medication prescribed? | Text |
| Adverse Event Reporting System | AERS | Adverse events | Adverse reactions | Keywords |
| AgingGenesDB | AGDB | Aging related genes | Keywords | Keywords |
| ClinicalTrials.gov | CT | Clinical trials | Condition | Keywords |
| DrugBank | DBK | Drugs | Drug category | Keywords |
| PharmGKB-Gene | PGGE | Relationships about drugs, diseases and genes | Gene related diseases | Keywords |
| UniProt KB | UPKB | Proteins | Biological processes | Keywords |

Figure 6 and Figure 7 show a representation of the coverage provided by both systems for each database and input type. Table 7 and Table 8 provide a summary of the evaluation results.

**Figure 6. Coverage distribution for the first ontology suggested by Ontology Recommender 1.0 (dashed red line) and 2.0 (solid blue line), using the individual ontologies output, for 600 texts extracted from 6 widely known databases (100 texts each)**. Vertical lines represent the mean coverage provided by the first ontology returned by Ontology Recommender 1.0 (dotted red line) and 2.0 (dashed-dotted blue line). The X-axis indicates the percentage of words covered by the ontology. The Y-axis displays the number of inputs for which a particular coverage percentage was obtained. AUTDB: Autism Database; GEO: Gene Expression Omnibus; GM: ARRS GoldMiner; IDV: Integrated Disease View; PM: PubMed; PMH: PubMed Health Drugs.

**Figure 7. Coverage distribution for the first ontology suggested by Ontology Recommender 1.0 (dashed red line) and 2.0 (solid blue line), using the individual ontologies output, for 600 lists of keywords extracted from 6 widely known databases (100 lists of keywords each).** Vertical lines represent the mean coverage provided by the first ontology returned by Ontology Recommender 1.0 (dotted red line) and 2.0 (dashed-dotted blue line). The X-axis indicates the percentage of input keywords covered by the ontology. The Y-axis displays the number of inputs for which a particular coverage percentage was obtained. AERS: Adverse Event Reporting System; AGDB: AgingGenesDB; CT: ClinicalTrials.gov; DBK: DrugBank; PGGE: PharmGKB-Gene; UPKB: UniProt KB.

**Table 7. Summary of evaluation results for text inputs.**

| Database | Mean length[a] | Executions with coverage < 20%[b] | | Mean coverage (top ranked ontology)[c] | | | Execution time (seconds) | | |
|---|---|---|---|---|---|---|---|---|---|
| | | 1.0[*] | 2.0[**] | 1.0[*] | 2.0[**] | 2.0 (sets)[***] | 1.0[*] | 2.0[**] | 2.0 (sets)[***] |
| AUTDB | 128.8 | 12.0% | 0.0% | 66.8% | 76.0% | 90.3% | 12.2 | 18.3 | 26.9 |
| GEO | 146.4 | 8.0% | 0.0% | 70.7% | 76.9% | 92.9% | 11.2 | 17.2 | 26.1 |
| GM | 55.7 | 48.0% | 0.0% | 46.6% | 82.6% | 94.8% | 9.6 | 12.3 | 15.2 |
| IDV | 150.2 | 28.0% | 0.0% | 50.6% | 71.8% | 89.3% | 9.5 | 13.1 | 21.1 |
| PM | 208.9 | 7.0% | 0.0% | 69.1% | 73.8% | 93.1% | 13.8 | 21.2 | 36.9 |
| PMH | 77.4 | 13.0% | 0.0% | 61.1% | 73.5% | 91.9% | 8.0 | 10.5 | 13.3 |
| **Mean** | **127.9** | **19.3%** | **0.0%** | **60.8%** | **75.7%** | **92.1%** | **10.7** | **15.4** | **23.2** |

[a] Mean of the number of words for the inputs extracted from the database.

[b] Percentage of executions where the coverage of the top recommended ontology was lower than 20%.

[c] Mean coverage provided by the top ranked ontology.

[*] Ontology Recommender 1.0; [**] Ontology Recommender 2.0; [***] Ontology Recommender 2.0 (ontology sets output).

**Table 8. Summary of evaluation results for keyword inputs.**

| Database | Mean length[a] | Executions with coverage < 20%[b] | | Mean coverage (top ranked ontology)[c] | | | Execution time (seconds) | | |
|---|---|---|---|---|---|---|---|---|---|
| | | 1.0[*] | 2.0[**] | 1.0[*] | 2.0[**] | 2.0 (sets)[***] | 1.0[*] | 2.0[**] | 2.0 (sets)[***] |
| AERS | 29.5 | 6.0% | 0.0% | 54.6% | 97.8% | 99.6% | 10.2 | 9.1 | 10.5 |
| AGDB | 8.2 | 5.0% | 0.0% | 53.4% | 67.5% | 82.9% | 6.5 | 9.9 | 10.9 |
| CT | 16.6 | 12.0% | 2.0% | 61.4% | 76.5% | 84.8% | 9.9 | 8.4 | 10.2 |
| DBK | 5.9 | 13.0% | 1.0% | 60.5% | 74.7% | 89.6% | 4.3 | 6.8 | 7.3 |
| PGGE | 15.4 | 2.0% | 0.0% | 73.1% | 80.5% | 83.0% | 9.9 | 9.1 | 10.3 |
| UPKB | 29.9 | 18.0% | 0.0% | 74.6% | 96.3% | 99.1% | 16.5 | 13.1 | 16.9 |
| **Mean** | **17.6** | **9.3%** | **0.5%** | **62.9%** | **82.2%** | **89.8%** | **9.5** | **9.4** | **11.0** |

[a] Mean of the number of words for the inputs extracted from the database.
[b] Percentage of executions where the coverage of the top recommended ontology was lower than 20%.
[c] Mean coverage provided by the top ranked ontology.
[*] Ontology Recommender 1.0; [**] Ontology Recommender 2.0; [***] Ontology Recommender 2.0 (ontology sets output).

For some inputs, the first ontology suggested by Ontology Recommender 1.0 provides very low coverage (under 20%). This results from one of the shortcomings previously described: Ontology Recommender 1.0 occasionally assigns a high score to ontologies that provide low coverage because they contain several classes matching the input. The new recommendation approach used by Ontology Recommender 2.0 addresses this problem: Virtually none of its executions provide such low coverage.

For example, Table 9 shows the ontologies recommended if we input the following description of a disease, extracted from the Integrated Disease View (IDV) database: *Chronic fatigue syndrome refers to severe, continued tiredness that is not relieved by rest and is not directly caused by other medical conditions. See also: Fatigue. The exact cause of chronic fatigue syndrome (CFS) is* *unknown. The following may also play a role in the development of CFS: CFS most commonly occurs in women ages 30 to 50.*

Ontology Recommender 1.0 suggests the Bone Dysplasia Ontology (BDO), whereas Ontology Recommender 2.0 suggests the NCI Thesaurus (NCIT). Because BDO covers only 4 of the input terms, while NCIT covers 17, the recommendation provided by Ontology Recommender 2.0 is more appropriate than that of its predecessor.

**Table 9. Comparison of the terms covered by Ontology Recommender 1.0 and Ontology Recommender 2.0 for the input text previously shown.**

| Ontology | Position | | Terms covered |
| --- | --- | --- | --- |
| | Ontology Recommender 1.0 | Ontology Recommender 2.0 | |
| BDO | 1 | 23 | *Chronic, severe, chronic, unknown* |
| NCIT | 2 | 1 | *Chronic fatigue syndrome, severe, continued, rest, directly, medical, Fatigue, exact, cause, chronic, fatigue syndrome, unknown, suggest, due to, following, role, development, ages* |

Ontology Recommender 2.0 also provides better mean coverage for both input types (i.e., text and keywords) across all the biomedical databases included in the evaluation. Compared to Ontology Recommender 1.0, the mean coverage reached using Ontology Recommender 2.0 was 14.9% higher for texts and 19.3% higher for keywords. That increase was even greater using the "ontology sets" output type provided by Ontology Recommender 2.0, which reached a mean coverage of 92.1% for texts (31.3% higher than the Ontology Recommender 1.0 ratings) and 89.8% for keywords (26.9% higher).

For the selected texts, the average execution time of Ontology Recommender 2.0 for the "ontologies" output is 15.4 seconds, 43.9% higher than the Ontology Recommender 1.0 execution time (10.7 seconds). The ontology recommendation process performed by Ontology Recommender 2.0 is much more complex than the one performed by the original version, and this is reflected by the execution times. The average execution time for keywords is similar in both systems (9.5 seconds for Ontology Recommender 1.0 and 9.4 seconds for Ontology Recommender 2.0). When dealing with keywords, the complex process performed by Ontology Recommender 2.0 is compensated by its ability to discard unnecessary annotations before staring the ontology evaluation process. These execution times are substantially better than those reported for similar systems. For example, the BiOSS system [21] needed an average of 207 seconds to process 30 keywords with a repository of 200 candidate ontologies. Performance of Ontology Recommender 2.0 is reasonable for general scenarios, where the quality of the suggestions is typically more important than the execution time.

## 3.2   Experiment 2: Refining Recommendations

Our second experiment set out to examine whether Ontology Recommender 2.0 is effective at discerning how to make meaningful recommendations when ontologies exhibit similar coverage of the input text. Specifically, we were interested in analyzing how the new version uses ontology acceptance, detail and specialization to prioritize the most appropriate ontologies.

We started with the 1200 inputs (600 texts and 600 lists of keywords) from the previous experiment, and selected those inputs for which the two versions of Ontology Recommender suggested different ontologies with similar coverage. We considered two coverage values similar if the difference between them was less than 10%. This yielded a total of 284 inputs (32 input texts and 252 lists of keywords). We executed both systems for those 284 inputs and analyzed the ontologies obtained in terms of their acceptance, detail and specialization scores.

Figure 8 and Table 10 show the results obtained. The ontologies suggested by Ontology Recommender 2.0 have higher acceptance (87.1) and detail scores (72.1) than those suggested by Ontology Recommender 1.0. Importantly, the graphs show peaks of low acceptance (<30%) and detail (<20%) for Ontology Recommender 1.0 that are addressed by Ontology Recommender 2.0.

The ontologies suggested by Ontology Recommender 2.0 have, on average, lower specialization scores (65.1) than those suggested by Ontology Recommender 1.0 (95.1). This is an expected result, given that the recommendation approach used by Ontology Recommender 1.0 is based on the relation between the number of annotations provided by each ontology and its size, which is our measure for ontology specialization.

Ontology Recommender 1.0 is better than Ontology Recommender 2.0 at finding small ontologies that provide multiple annotations for the user's input. However, those ontologies are not necessarily the most appropriate to describe the input data. As we have seen (see Section 1.2.1), a large number of annotations does not always indicate a high input coverage. Ontology Recommender 1.0 sometimes suggests ontologies with high specialization scores but with very low input coverage, which makes the ontologies inappropriate for the user's input. The multi-criteria evaluation approach used by Ontology Recommender 2.0 has been designed to address this issue by evaluating ontology specialization in combination with other criteria, including ontology coverage.

**Figure 8. Acceptance, detail and specialization distribution for the first ontology suggested by Ontology Recommender 1.0 (dashed red line) and 2.0 (solid blue line), for the 284 inputs selected.** Vertical lines represent the mean acceptance, detail and specialization scores provided by Ontology Recommender 1.0 (dotted red line) and 2.0 (dashed-dotted blue line). The X-axis indicates the acceptance, detail and specialization score provided by the top ranked ontology. The Y-axis displays the number of inputs for which a particular score was obtained.

**Table 10. Mean acceptance, detail and specialization scores provided by the two versions of Ontology Recommender for experiment 2.**

|  | Text (32 inputs) | | Keywords (252 inputs) | | All (284 inputs) | |
|---|---|---|---|---|---|---|
|  | 1.0[*] | 2.0[**] | 1.0[*] | 2.0[**] | 1.0[*] | 2.0[**] |
| Mean acceptance | 91.3 | 99.2 | 39.8 | 85.2 | **45.7** | **87.1** |
| Mean detail | 5.8 | 56.1 | 15.7 | 73.9 | **14.6** | **72.1** |
| Mean specialization | 94.7 | 90.3 | 94.8 | 61.6 | **95.1** | **65.1** |

[*]Ontology Recommender 1.0; [**]Ontology Recommender 2.0.

## 3.3 Experiment 3: High Coverage and Specialized Ontologies

We set out to evaluate how well Ontology Recommender 2.0 prioritizes recommending small ontologies that provide appropriate coverage for the input data. We created 15 inputs, each of which contained keywords from a very specific domain (e.g., adverse reactions, dermatology, units of measurement), and executed both versions of the Ontology Recommender for those inputs.

Table 11 shows the particular domain for each of the 15 inputs used, and the first ontology suggested by each version of Ontology Recommender, as well as the size of each ontology and the coverage provided.

Analysis of the results reveals that Ontology Recommender 2.0 is more effective than Ontology Recommender 1.0 for suggesting specialized ontologies that provide high input coverage. In 9 out of 15 inputs (60%), the first ontology suggested by Ontology Recommender 2.0 is more appropriate, in terms of its size and coverage provided, than the ontology recommended by Ontology Recommender 1.0. Ontology Recommender 2.0 considers input coverage in addition to ontology specialization, which Ontology Recommender 1.0 does not. In addition, Ontology Recommender 2.0 uses a different annotation scoring method (the function *annotationScore2(a)*; see Section 2.1.1) that gives more weight to annotations that cover multi-word terms. There is one input (no. 13), for which the ontology suggested by Ontology Recommender 2.0 provides higher coverage (88% versus 80%), but it is bigger than the ontology recommended by Ontology Recommender 1.0 (324K classes versus 119K). In 5 out of 15 inputs (33%), both systems recommended the same ontology.

**Table 11. Experiment 3 results.** The table shows the input size (number of keywords) and domain, as well as the first ontology suggested by Ontology Recommender 1.0 and Ontology Recommender 2.0. The size of each ontology (number of classes) and the coverage provided are also shown. The best results for each input (lowest ontology size and highest coverage), are highlighted in bold.

| Input | Size | Input domain | Ontology Recommender 1.0 | | | Ontology Recommender 2.0 | | |
|---|---|---|---|---|---|---|---|---|
| | | | Result | Size | Coverage (%) | Result | Size | Coverage (%) |
| 1 | 23 | Adverse reactions | SNOMEDCT | 324,129 | 52.1 | MEDDRA | **68,261** | **91.3** |
| 2 | 8 | Autism spectrum disorder | ASDPTO | 284 | 100.0 | ASDPTO | 284 | 100.0 |
| 3 | 14 | Bioinformatics operations | SWO | 4,068 | 100.0 | EDAM | **3,240** | 100.0 |
| 4 | 30 | Biomedical investigations | NCIT | 118,941 | 63.3 | OBI | **3,055** | **100.0** |
| 5 | 26 | Cell types | SYN | 14,462 | 76.9 | CL | **6,532** | **88.4** |
| 6 | 14 | Clinical research | NCIT | 118,941 | 78.6 | OCRE | **389** | **100.0** |
| 7 | 13 | Dermatology | DERMLEX | 6,106 | 92.3 | DERMLEX | 6,106 | 92.3 |
| 8 | 14 | Environmental features | ENVO | 2,307 | 100.0 | ENVO | 2,307 | 100.0 |
| 9 | 18 | Enzyme sources | NCIT | 118,941 | 55.6 | BTO | **5,902** | **72.2** |
| 10 | 19 | Fish anatomy | NIFSTD | 124,337 | 68.4 | TAO | **3,428** | **79.0** |
| 11 | 44 | Human diseases | RH-MESH | 305,349 | 52.3 | DOID | **11,280** | **95.5** |
| 12 | 13 | Mathematical models in life sciences | MAMO | 100 | 100.0 | MAMO | 100 | 100.0 |
| 13 | 25 | Primary care | NCIT | **118,941** | 80.0 | SNOMEDCT | 324,129 | **88.0** |
| 14 | 23 | Signs and symptoms | NCIT | 118,941 | 65.2 | SYMP | **936** | **91.3** |
| 15 | 25 | Units of Measurement | TEO | 687 | 92.0 | TEO | 687 | 92.0 |

# 4   Discussion

Recommending biomedical ontologies is a challenging task. The great number, size, and complexity of biomedical ontologies, as well as the diversity of user requirements and expectations, make it difficult to identify the most appropriate ontologies to annotate biomedical data. The analysis of the results demonstrates that ontologies suggested using our new recommendation approach are more appropriate than those recommended using the original method. Our acceptance evaluation method has proved to be successful to rank ontologies, and it is currently used not only by the Ontology Recommender, but also by the BioPortal search engine. The classes returned when searching in BioPortal are ordered according to the general acceptance of the ontologies to which they belong.

We note that, because the system is designed in a modular way, it will be easy to add new evaluation criteria to extend its functionality. As a first priority, we intend to improve and extend the evaluation criteria currently used. In addition, we will investigate the effect of extending the Ontology Recommender to include relevant features not yet considered, such as the frequency of an ontology's updates, its levels of abstraction, formality, granularity, and the language in which the ontology is expressed.

Indeed, using metadata information is a simple but often ignored approach to select ontologies. Coverage-based approaches often miss relevant results because they focus on the content of ontologies and ignore more general information about the ontology. For example, applying the new Ontology Recommender to the Wikipedia definition of anatomy[13] will return some widely-known ontologies that contain the terms *anatomy*, *structure*, *organism* and *biology*, but the Foundational Model of Anatomy (FMA), which is the reference ontology about human anatomy will not show up in the top 25 results. Our specialization criterion uses the content of the ontology and the ontology size to discriminate between large ontologies and small ontologies that have better specialization. However, ontologies that provide multiple annotations for the input data are not always specialized to deal with the input domain. Sometimes very specialized ontologies for a domain may provide low coverage for a particular text from the domain. In this scenario, metadata about the domain of the ontology (e.g., 'anatomy' in the case of FMA) could be used to enhance our ontology specialization criterion by limiting the suggestions to those ontologies whose domain matches the input data domain. We are currently refining, in collaboration with the Center for Expanded Data Annotation and Retrieval (CEDAR) [41] and the AgroPortal ontology repository [42], the way BioPortal handles metadata for ontologies in order to support even more ontology recommendation scenarios.

---

[13] https://en.wikipedia.org/wiki/Anatomy

Our coverage evaluation approach may be further enhanced by complementing our annotation scoring method (i.e., *annotationScore2*) with term extraction techniques. We plan to analyze the application of a term extraction measure, called C-value [43], which is specialized for multi-word term extraction, and that has already been applied to the results of the NCBO Annotator, leading to significant improvements [44].

There are some possible avenues for enhancing our assessment of ontology acceptance. These include considering the number of projects that use a specific ontology, the number of mappings created manually that point to a particular ontology, the number of user contributions (e.g., mappings, notes, comments), the metadata available per ontology, and the number, publication date and publication frequency of ontology versions. There are other indicators external to BioPortal that could be useful for performing a more comprehensive evaluation of ontology acceptance, such as the number of Google results when searching for the ontology name or the number of PubMed publications that contain the ontology name [21].

Reusing existing ontologies instead of building new ones from scratch has many benefits, including lowering the time and cost of development, and avoiding duplicate efforts [45]. As shown by a recent study [46], reuse is fairly low in BioPortal, but there are some ontologies that are approaching complete reuse (e.g., Mental Functioning Ontology). Our approach should be able to identify these ontologies and assign them a lower score than those ontologies where the knowledge was first defined. We will study the inclusion of additional evaluation criteria to weigh the amount of original knowledge provided by a particular ontology for the input data.

The current version of Ontology Recommender uses a set of default parameters to control how the different evaluation scores are calculated, weighted and aggregated. These parameters provide acceptable results for general ontology recommendation scenarios, but some users may need to modify the default settings to match their needs. In the future, we would like the system to use an automatic weight adjustment approach. We will investigate whether it is possible to develop methods of adjusting the weights dynamically for specific scenarios.

Ontology Recommender helps to identify all the ontologies that would be suitable for semantic annotation. However, given the number of ontologies in BioPortal, it would be difficult, computationally expensive, and often useless to annotate user inputs with all the ontologies in the repository. Ontology Recommender could function within BioPortal as a means to screen ontologies for use with the NCBO Annotator. Note that the output of the Annotator is a ranked list of annotations performed with multiple ontologies, while the output of the Ontology Recommender is a ranked list of ontologies. A user might be offered the possibility to "Run the Ontology Recommender first" before actually calling the Annotator. Then only the top-ranked ontologies would be used for annotations.

A user-based evaluation would help us understand the system's utility in real-world settings. Our experience evaluating the original Ontology Recommender and BiOSS showed us that obtaining a user-based evaluation of an ontology recommender system is a challenging task. For example, the evaluators of BiOSS reported that they would need at least 50 minutes to perform a high-quality evaluation of the system for each test case. We plan to investigate whether crowd-sourcing methods, as an alternative, can be useful to evaluate ontology recommendation systems from a user-centered perspective.

Our approach for ontology recommendation was designed for the biomedical field, but it can be adapted to work with ontologies from other domains so long as they have a resource equivalent to the NCBO Annotator, an API to obtain basic information about all the candidate ontologies, and their classes, and alternative resources for extracting information about the acceptance of each ontology. For example, AgroPortal [42] is an ontology repository based on NCBO BioPortal technology. AgroPortal uses Ontology Recommender 2.0 in the context of plant, agronomic and environmental sciences.[14]

# 5 Conclusions

Biomedical ontologies are crucial for representing knowledge and annotating data. However, the large number, complexity, and variety of biomedical ontologies make it difficult for researchers to select the most appropriate ontologies for annotating their data. In this paper, we presented a novel approach for recommending biomedical ontologies. This approach has been implemented as release 2.0 of the NCBO Ontology Recommender, a system that is able to find the best ontologies for a biomedical text or set of keywords. Ontology Recommender 2.0 combines the strengths of its predecessor with a range of adjustments and new features that improve its reliability and usefulness. Our evaluation shows that, on average, the new system is able to suggest ontologies that provide better input coverage, contain more detailed information, are more specialized, and are more widely accepted than those suggested by the original Ontology Recommender. In addition, the new version is able to evaluate not only individual ontologies, but also different ontology sets, in order to maximize input coverage. The new system can be customized to specific user needs and it provides more explanatory output information than its predecessor, helping users to understand the results returned. The new service, embedded into the NCBO BioPortal, will be a more valuable resource to the community of researchers, scientists, and developers working with ontologies.

---

[14] http://agroportal.lirmm.fr/recommender

## List of abbreviations

BioPortal ontology acronyms:

| BIOMODELS | BioModels Ontology (BIOMODELS) |
|---|---|
| CPT | Current Procedural Terminology |
| EFO | Experimental Factor Ontology |
| EHDA | Human Developmental Anatomy Ontology, timed version |
| EP | Cardiac Electrophysiology Ontology |
| FMA | Foundational Model of Anatomy |
| HUPSON | Human Physiology Simulation Ontology |
| LOINC | Logical Observation Identifier Names and Codes |
| MEDDRA | Medical Dictionary for Regulatory Activities |
| MP | Mammalian Phenotype Ontology |
| NCIT | National Cancer Institute Thesaurus |
| NDDF | National Drug Data File |
| NDFRT | National Drug File - Reference Terminology |
| RXNORM | RxNORM |
| SNOMEDCT | Systematized Nomenclature of Medicine - Clinical Terms |
| SWEET | Semantic Web for Earth and Environment Technology Ontology |
| VSO | Vital Sign Ontology |
| MESH | Medical Subject Headings |
| ICD9CM | International Classification of Diseases, Version 9 - Clinical Modification |

| RCD | Read Codes, Clinical Terms Version 3 |
| OMIM | Online Mendelian Inheritance in Man |
| VANDF | Veterans Health Administration National Drug File |
| GO | Gene Ontology |
| CRISP | Computer Retrieval of Information on Scientific Projects Thesaurus |
| ICPC | International Classification of Primary Care |
| MEDLINEPLUS | MedlinePlus Health Topics |
| COSTART | Coding Symbols for a Thesaurus of Adverse Reaction Terms |
| PDQ | Physician Data Query |
| SYMP | Symptom Ontology |

# Declarations

### Ethics approval and consent to participate

Not applicable.

### Consent for publication

Not applicable.

### Availability of data and materials

- Project name: The Biomedical Ontology Recommender.

- Project home page: http://bioportal.bioontology.org/recommender.

- Project GitHub repository: https://github.com/ncbo/ncbo_ontology_recommender.

- REST service parameters: http://data.bioontology.org/documentation#nav_recommender.

- Operating system(s): Platform independent.

- Programming language: Ruby, Javascript, HTML.

- Other requirements: none.
- License: BSD (http://www.bioontology.org/BSD-license).
- Datasets used in our evaluation: https://git.io/vDIXV.

## Competing interests

The authors declare no competing interests.

## Funding

This work was supported in part by the National Center for Biomedical Ontology as one of the National Centers for Biomedical Computing, supported by the NHGRI, the NHLBI, and the NIH Common Fund under grant U54 HG004028 from the U.S. National Institutes of Health. Additional support was provided by CEDAR, the Center for Expanded Data Annotation and Retrieval (U54 AI117925) awarded by the National Institute of Allergy and Infectious Diseases through funds provided by the trans-NIH Big Data to Knowledge (BD2K) initiative. This project has also received support from the European Union's Horizon 2020 research and innovation programme under the Marie Sklodowska-Curie grant agreement No 701771 and the French National Research Agency (grant ANR-12-JS02-01001).

## Author's contributions

MMR conceived the approach, designed and implemented the system, and drafted the initial manuscript. CJ participated in technical discussions and provided ideas to refine the approach. MAM supervised the work and gave advice and feedback at all stages. CJ, MJO, JG, and AP provided critical revision and edited the manuscript. All authors gave the final approval of the manuscript.

**Acknowledgments**

The authors acknowledge the suggestions about the problem of recommending ontologies provided by the NCBO team, as well as their assistance and advice on integrating Ontology Recommender 2.0 into BioPortal. The authors also thank Simon Walk for his report on the BioPortal traffic data. Natasha Noy and Vanessa Aguiar offered valuable feedback.

**Endnotes**

[1] The BioPortal API received 18.7M calls/month on average in 2016. The BioPortal website received 306.8K pageviews/month on average in 2016 (see Additional file 1 for more detailed traffic data). The two main BioPortal papers [3, 4] accumulate 923 citations at the time of writing this paper, with 145 citations received in 2016.

[2] http://bioportal.bioontology.org/

[3] At the time of writing this paper, there are 6357 citations to the NCBO Ontology Recommender 1.0 paper [6]. The Ontology Recommender 1.0 serviceAPI received has received more than 7.1K95 calls per month on average in 2014. Detailed traffic data is provided in Additional file 1.

[4] This formula is slightly different from the scoring method presented in the paper describing the original Ontology Recommender Web service [5]. It corresponds to an upgrade done in the recommendation algorithm in December 2011, when BioPortal 3.5 was released, for which description and methodology was never published. The normalization strategy was improved by applying a logarithmic transformation to the ontology size to avoid a negative effect on very large ontologies. Mappings between ontologies, used to favor reference ontologies, were discarded due to the small number of manually created and curated mappings that could be used for such a purpose. The hierarchy-based semantic expansion was replaced by the position of the matched class in the ontology hierarchy.

[5] See "List of abbreviations".

[6] The function is called *annotationScore2* to differentiate it from the original *annotationScore* function.

[7] The API documentation is available at http://data.bioontology.org/documentation#nav_recommender

[8] The Web-based user interface is available at http://bioportal.bioontology.org/recommender

[9] https://www.bioontology.org/wiki/index.php/Category:NCBO_Virtual_Appliance

[10] BioPortal release notes: https://www.bioontology.org/wiki/index.php/BioPortal_Release_Notes

[11] https://github.com/ncbo/ncbo_ontology_recommender.

[12] The NCBO Resource Index is an ontology-based index that provides access to over 30 million biomedical records from 48 widely-known databases. It is available at: http://bioportal.bioontology.org.

[13] https://en.wikipedia.org/wiki/Anatomy.

[14] http://agroportal.lirmm.fr/recommender.

**Additional files**

Additional file 1:

- Format: Word document (.docx).
- Title of data: Ontology Recommender traffic summary.
- Description of data: Summary of traffic received by the Ontology Recommender for the period 2014-2016, compared to the other most used BioPortal services.

Additional file 2:

- Format: Word document (.docx).
- Title of data: Default configuration settings.
- Description of data: Default values used by the NCBO Ontology Recommender 2.0 for the parameters that control how the different scores are calculated, weighted and aggregated.

| BioPortal API Traffic Summary[*] | | | | | | | | |
|---|---|---|---|---|---|---|---|---|
| **Service** | **2014** | | **2015** | | | **2016**[**] | | |
| | calls/mo[a] | %total[b] | calls/mo[a] | %total[b] | %var[c] | calls/mo[a] | %total[b] | %var[c] |
| **Annotator** | 2,390,392 | 22.83% | 1,961,791 | 13.99% | -17.93% | 484,795 | 2.58% | -75.29% |
| **Search** | 820,073 | 7.83% | 1,345,884 | 9.60% | 64.12% | 873,954 | 4.65% | -35.06% |
| **Mappings** | 285,846 | 2.73% | 502,298 | 3.58% | 75.72% | 213,003 | 1.13% | -57.59% |
| **Ontology Recommender** | **7,107** | **0.07%** | **6,359** | **0.05%** | **-10.53%** | **45,211** | **0.24%** | **611.00%** |
| **Resource Index** | 2,388 | 0.02% | 21,134 | 0.15% | 785.20% | 1,013 | 0.01% | -95.20% |
| **Other**[d] | 6,962,751 | 66.51% | 10,181,773 | 72.63% | 46.23% | 17,161,146 | 91.38% | 68.55% |
| **TOTAL calls/mo**[e] | 10,468,557 | 100.00% | 14,019,239 | 100.00% | 33.92% | 18,779,122 | 100.00% | 33.95% |

[*] The calls made from the BioPortal website to the BioPortal API have been excluded.
[**] 2016 data are based on the period Jan-Oct (10 months).
[a] Mean number of API calls/month.
[b] Percentage of API calls/month with respect to the total number of BioPortal API calls/month.
[c] Percentage of variation with respect to the previous year.
[d] Other requests, which include browsing ontology classes (class details, paths to root, class tree, children, parents, etc.), ontologies (ontology details, root classes, groups, submissions, metrics, analytics, etc.), instances, projects, and users.
[e] Total number of API calls/month for the BioPortal API.

| BioPortal Website Traffic Summary[*] | | | | | | | | |
|---|---|---|---|---|---|---|---|---|
| **Webpage** | **2014** | | **2015**[**] | | | **2016**[***] | | |
| | pv/mo[a] | %total[b] | pv/mo[a] | %total[b] | %var[c] | pv/mo[a] | %total[b] | %var[c] |
| **Annotator** | 3,371 | 0.79% | 2,687 | 0.90% | -20.30% | 3,048 | 0.99% | 13.44% |
| **Search** | 52,119 | 12.29% | 56,905 | 18.96% | 9.18% | 72,951 | 23.77% | 28.20% |
| **Mappings** | 2,789 | 0.66% | 7,213 | 2.40% | 158.64% | 2,258 | 0.74% | -68.70% |
| **Ontology Recommender** | **1,388** | **0.33%** | **925** | **0.31%** | **-33.38%** | **1,244** | **0.41%** | **34.55%** |
| **Resource Index** | 1,715 | 0.40% | 1,033 | 0.34% | -39.77% | 1,154 | 0.38% | 11.74% |
| **Other**[d] | 362,637 | 85.52% | 231,403 | 77.09% | -36.19% | 226,229 | 73.72% | -2.24% |
| **TOTAL pv/mo**[e] | 424,019 | 100.00% | 300,166 | 100.00% | -29.21% | 306,884 | 100.00% | 2.24% |

[*] Filtered traffic (excluding bots and AJAX requests).
[**] 2015 data are based on the periods Jan-Jul and Nov-Dec (9 months).
[***] 2016 data are based on the period Jan-Jul (7 months).
[a] Mean number of pageviews/month.
[b] Percentage of pageviews/month with respect to the total number of pageviews /month to the BioPortal website.
[c] Percentage of variation with respect to the previous year.
[d] Other pageviews, which include browsing ontology classes (class details, paths to root, class tree, children, parents, etc.), ontologies (ontology details, root classes, groups, submissions, metrics, analytics, etc.), instances, projects, and users.
[e] Total number of pageviews/month to the BioPortal website.

# Ontology Recommender 2.0 - Default configuration settings

| Ontology evaluation and score aggregation |||| 
|---|---|---|---|
| **Criterion** | **Weight** | **Other parameters** ||
| Ontology coverage | 0.55 | Preferred name (PREF) score: 10<br>Synonym (SYN) score: 5<br>Multi-word score: 3 ||
| Ontology acceptance | 0.15 | Presence | Weight ($w_{visits}$): 0.5<br>Repositories: UMLS |
| | | Visits | Weight ($w_{presence}$): 0.5<br>Repositories: BioPortal<br>Period: 12 months |
| Ontology detail | 0.15 | Definitions threshold ($k_d$): 1<br>Synonyms threshold ($k_s$): 3<br>Properties threshold ($k_p$): 17 ||
| Ontology specialization | 0.15 | N/A ||
| **Ontology ranking** ||||
| Ranking size | 25 |||
| Maximum number of ontologies/set | 3 |||

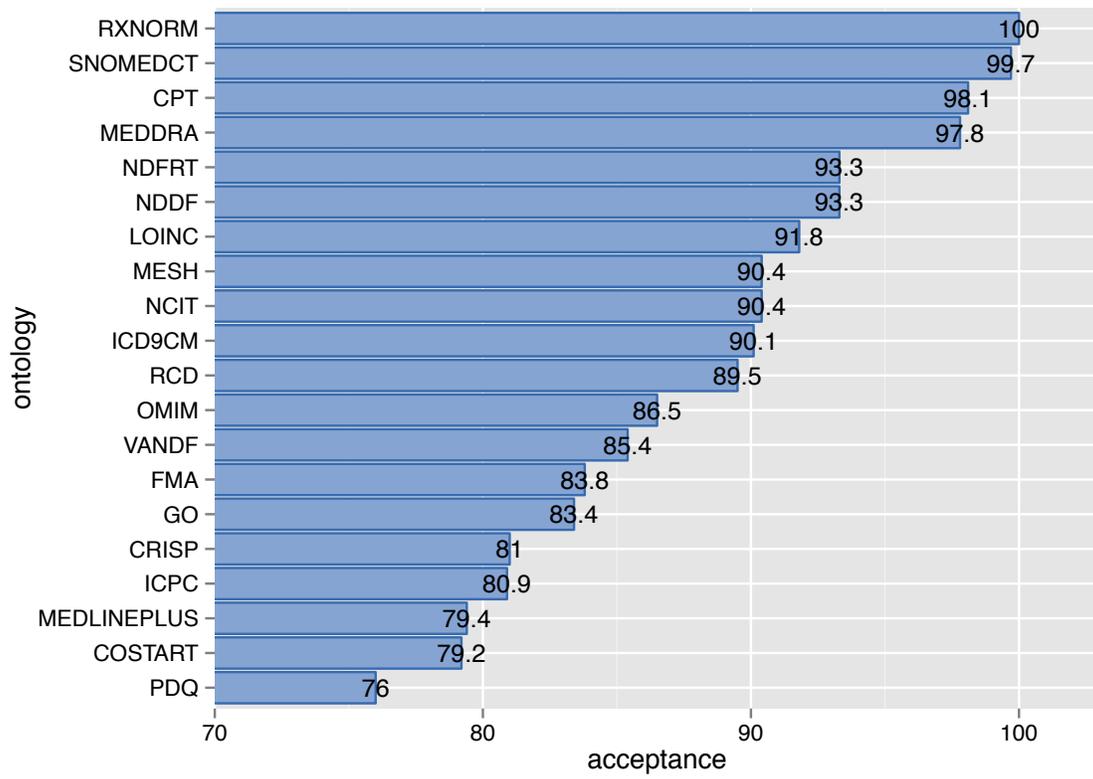

**Figure 1**

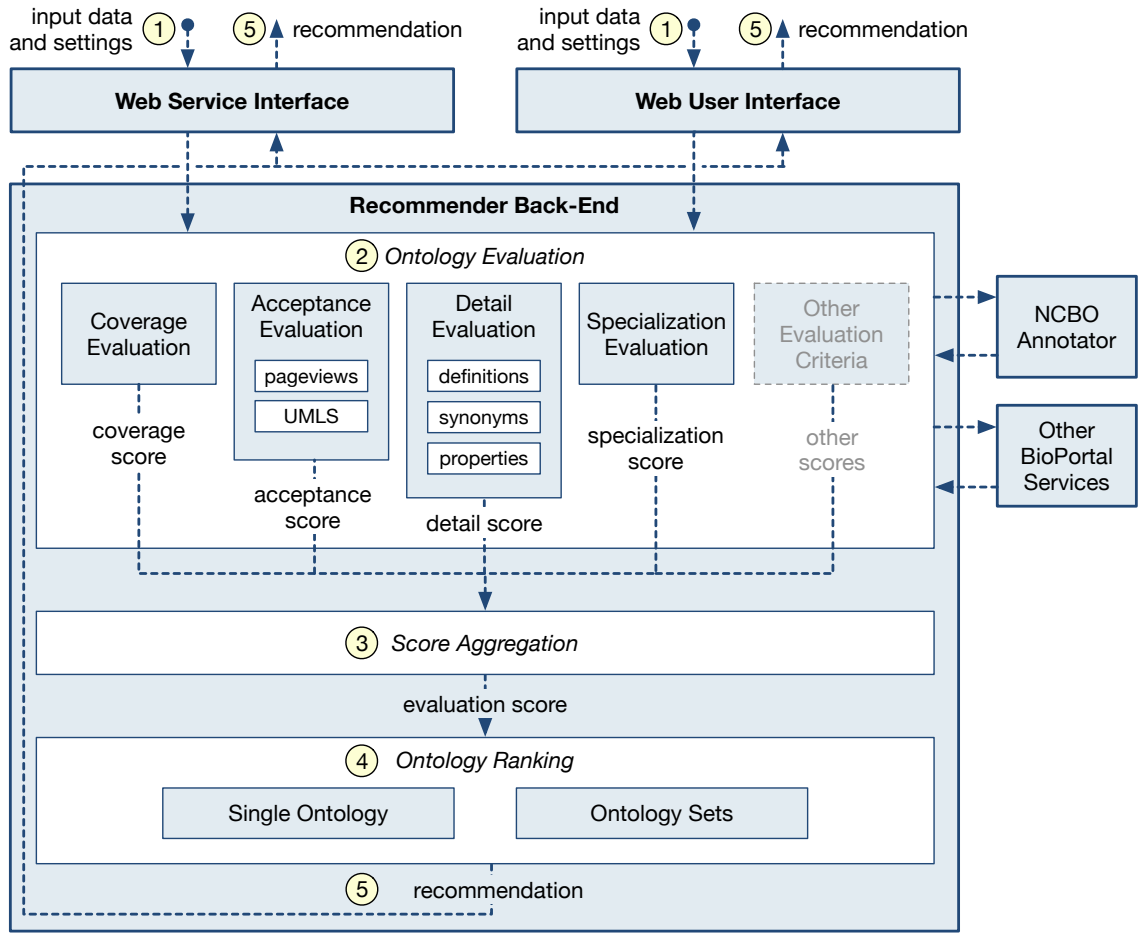

Figure 2

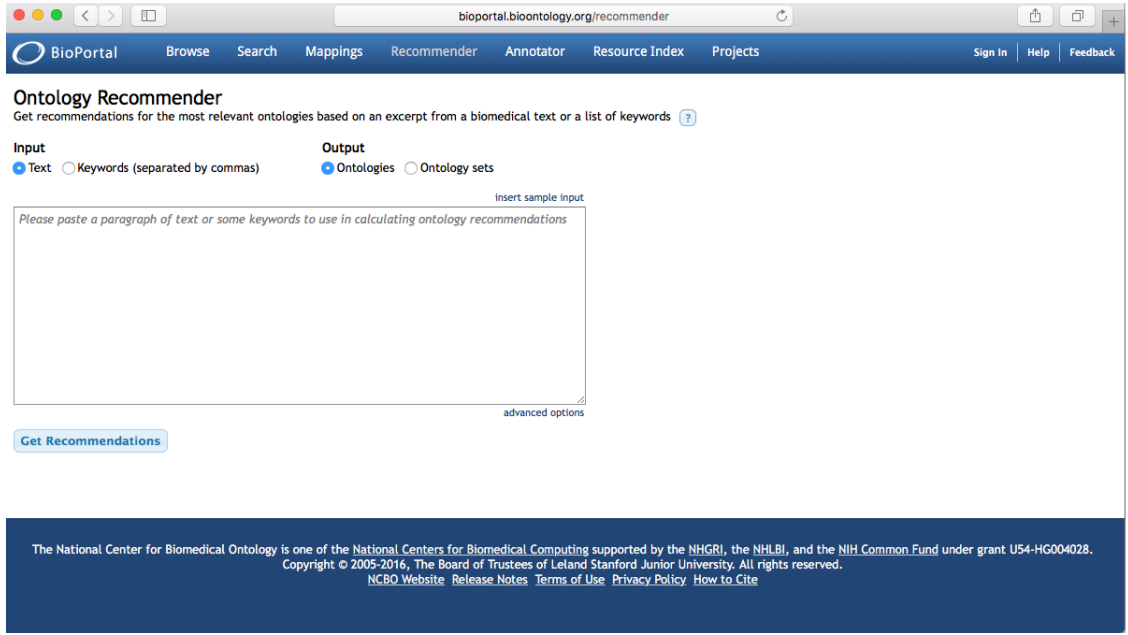

**Figure 3**

**Input**
○ Text  ● Keywords (separated by commas)

**Output**
● Ontologies  ○ Ontology sets

insert sample input

headaches, anemia, abnormal behavior, irritability, floppy head, cellulitis, lameness, stiff neck, facial tremor, backache, abdominal pain, weight gain, congestion, sneezing, respiratory failure, vascular alteration, atrial fibrillation, sleepy, sweaty, tired, weak

advanced options

[Edit Input]

**Recommended ontologies**

| POS. | ONTOLOGY | FINAL SCORE | COVERAGE SCORE | ACCEPTANCE SCORE | DETAIL SCORE | SPECIALIZATION SCORE | ANNOTATIONS | HIGHLIGHT ANNOTATIONS |
|---|---|---|---|---|---|---|---|---|
| 1 | SYMP | 74.3 | 90.2 | 29.1 | 36.3 | 99.2 | 17 | ☑ |
| 2 | SNOMEDCT | 68.9 | 67.9 | 95.3 | 59.2 | 55.9 | 16 | ☐ |
| 3 | NCIT | 64.0 | 55.1 | 87.6 | 74.9 | 62.3 | 13 | ☐ |
| 4 | MEDDRA | 60.3 | 64.2 | 96.5 | 28.5 | 41.3 | 14 | ☐ |
| 5 | MESH | 52.5 | 38.7 | 88.2 | 92.6 | 27.4 | 9 | ☐ |
| 6 | RCD | 51.5 | 50.2 | 86.7 | 34.5 | 38.0 | 10 | ☐ |
| 7 | CSSO | 47.6 | 43.6 | 22.6 | 77.9 | 56.8 | 9 | ☐ |

**Figure 4**

**Figure 5**

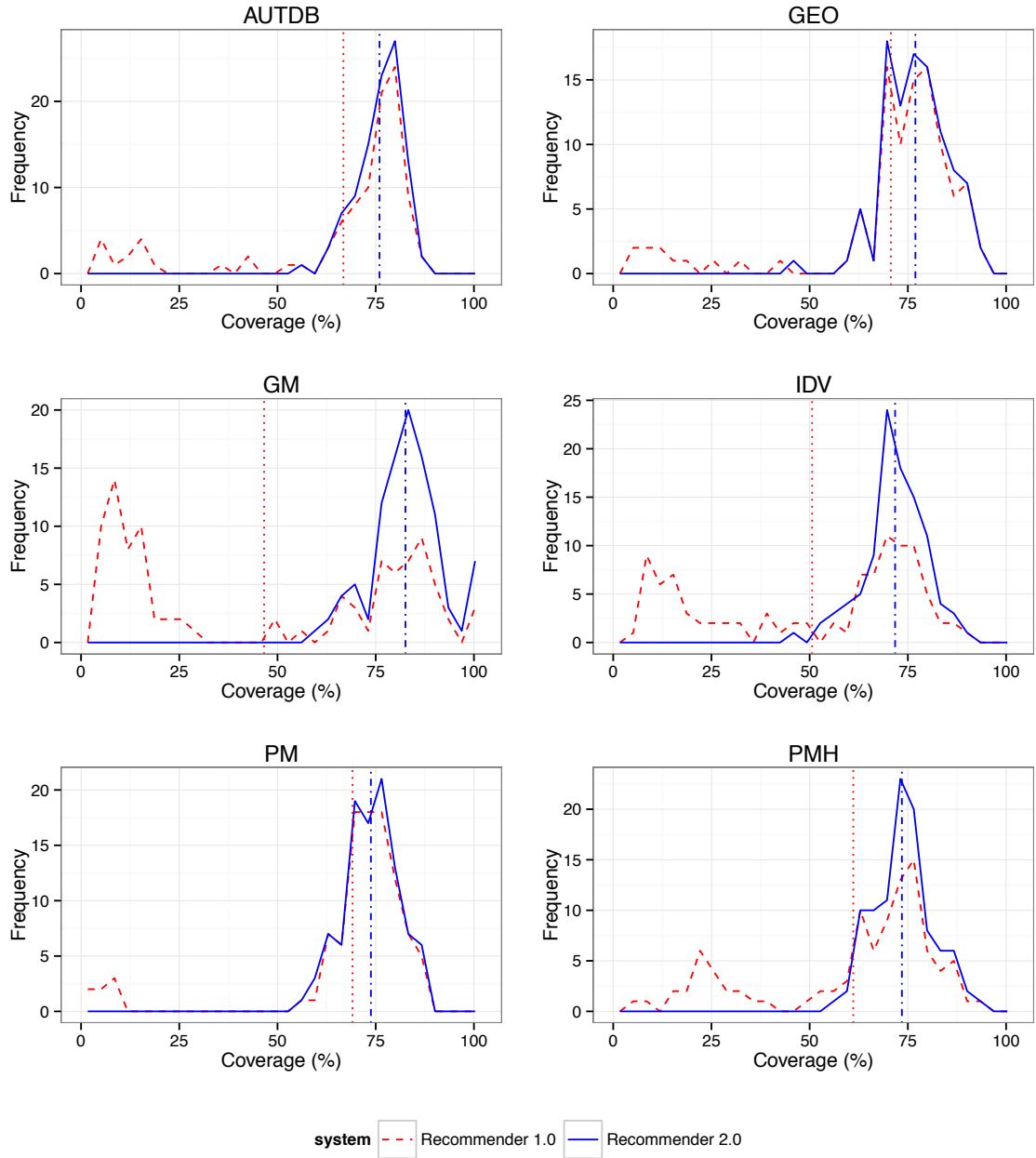

Figure 6

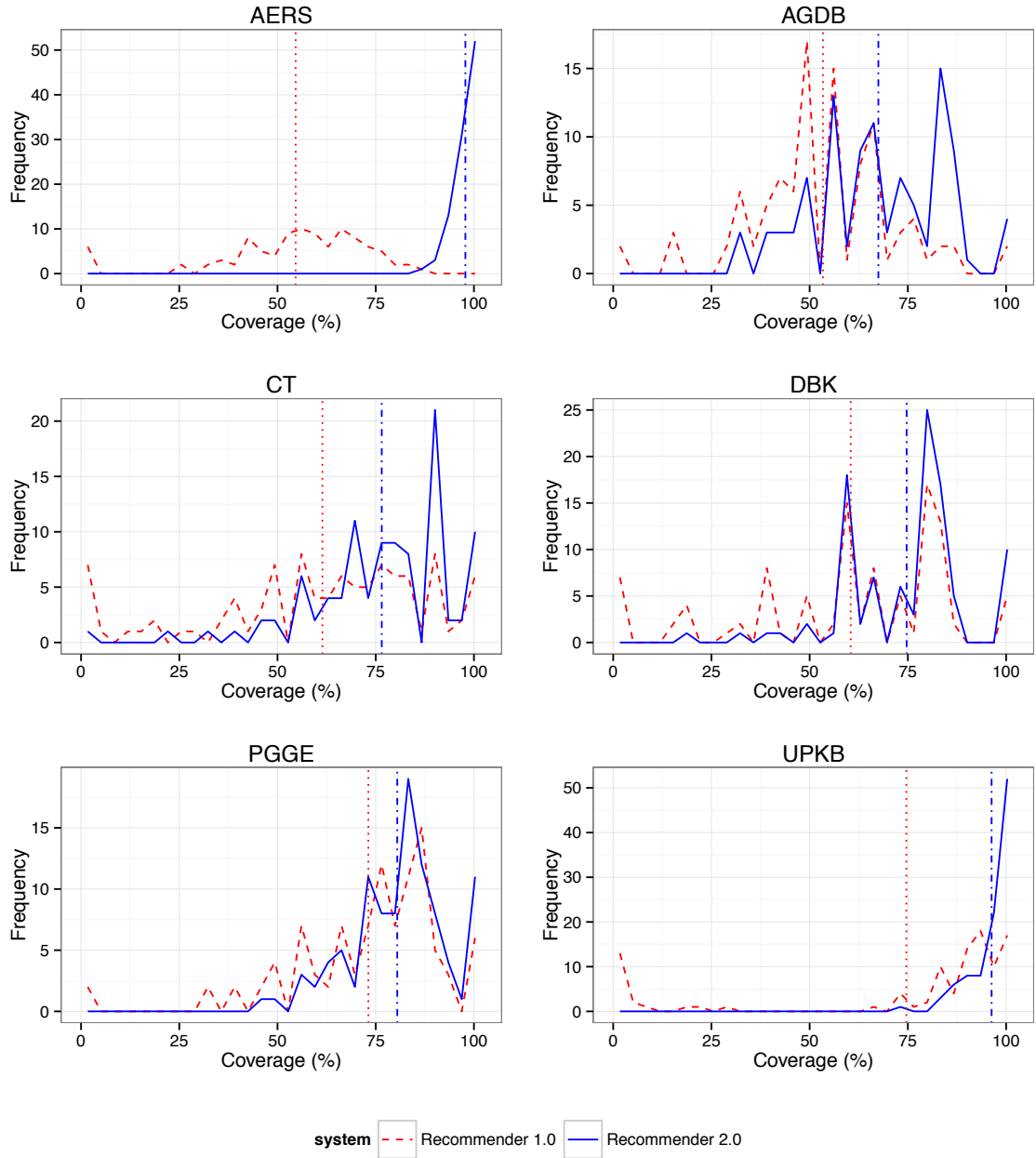

**Figure 7**

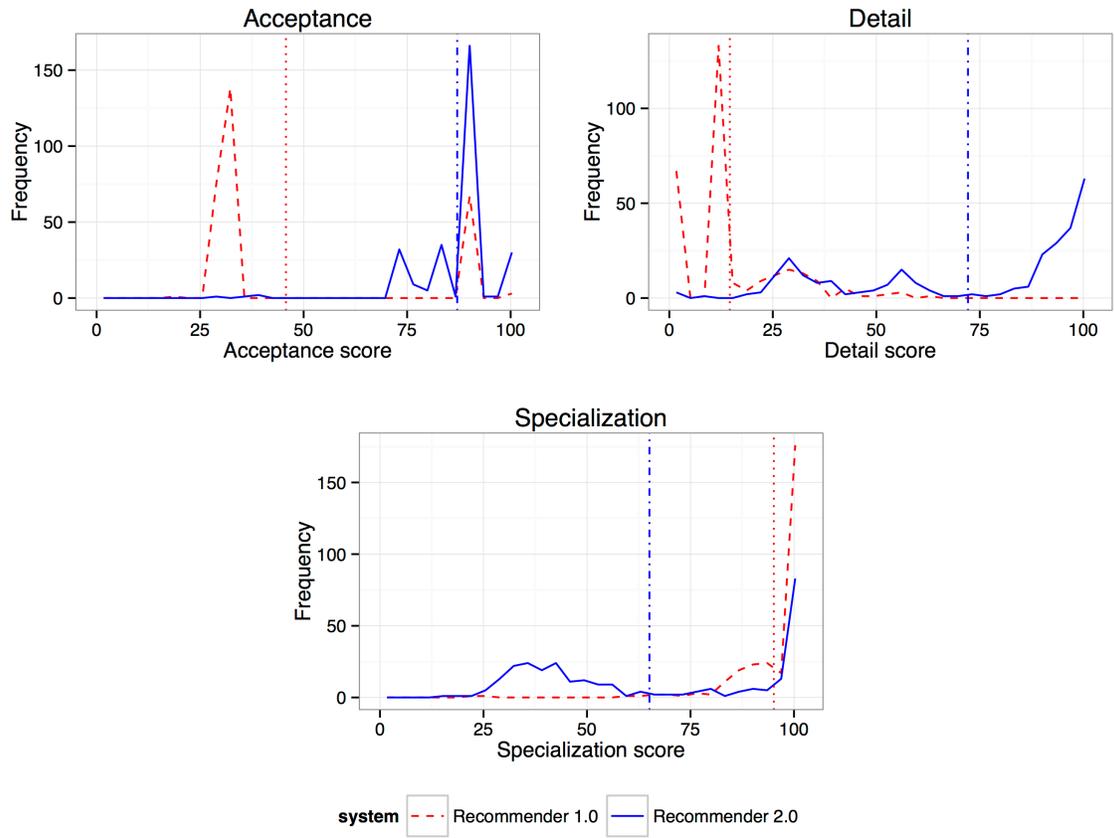

Figure 8